%% file: main.tex
\documentclass[letterpaper]{article} 
\usepackage[draft]{aaai25}  
\usepackage{times}  
\usepackage{helvet}  
\usepackage{courier}  
\usepackage[hyphens]{url}  
\usepackage{graphicx} 
\usepackage{natbib}  
\usepackage{caption} 
\usepackage{algorithm}
\usepackage{algorithmic}

\usepackage{booktabs}
\usepackage{amsmath}
\usepackage{xcolor}
\usepackage{subcaption}

\usepackage{newfloat}
\usepackage{listings}
\DeclareCaptionStyle{ruled}{labelfont=normalfont,labelsep=colon,strut=off} 
\floatstyle{ruled}
\newfloat{listing}{tb}{lst}{}
\floatname{listing}{Listing}
\title{VLM-KD: Knowledge Distillation from VLM for Long-Tail Visual Recognition}
\author{
    Zaiwei Zhang, Gregory P. Meyer, Zhichao Lu, \\ Ashish Shrivastava, Avinash Ravichandran, Eric M. Wolff
}
\affiliations{
    Cruise LLC\\

    333 Brannan St\\
    San Francisco, CA 94107, USA\\
}

\begin{document}

\maketitle

\begin{abstract}
  For visual recognition, knowledge distillation typically involves transferring knowledge from a large, well-trained teacher model to a smaller student model. In this paper, we introduce an effective method to distill knowledge from an off-the-shelf vision-language model (VLM), demonstrating that it provides novel supervision in addition to those from a conventional vision-only teacher model. Our key technical contribution is the development of a framework that generates novel text supervision and distills free-form text into a vision encoder. We showcase the effectiveness of our approach, termed VLM-KD, across various benchmark datasets, showing that it surpasses several state-of-the-art long-tail visual classifiers. To our knowledge, this work is the first to utilize knowledge distillation with text supervision generated by an off-the-shelf VLM and apply it to vanilla randomly initialized vision encoders.
\end{abstract}

%
\input{01_intro}
\input{02_related}
\input{03_approach}

\input{04_result}
\input{05_ablation}
\input{06_conclusion}


\input{main.bbl}
\clearpage
\appendix
\input{08_appendix}

\begin{figure*}[t!]
 \begin{subfigure}{0.49\linewidth}
         \centering
         \includegraphics[width=\textwidth]{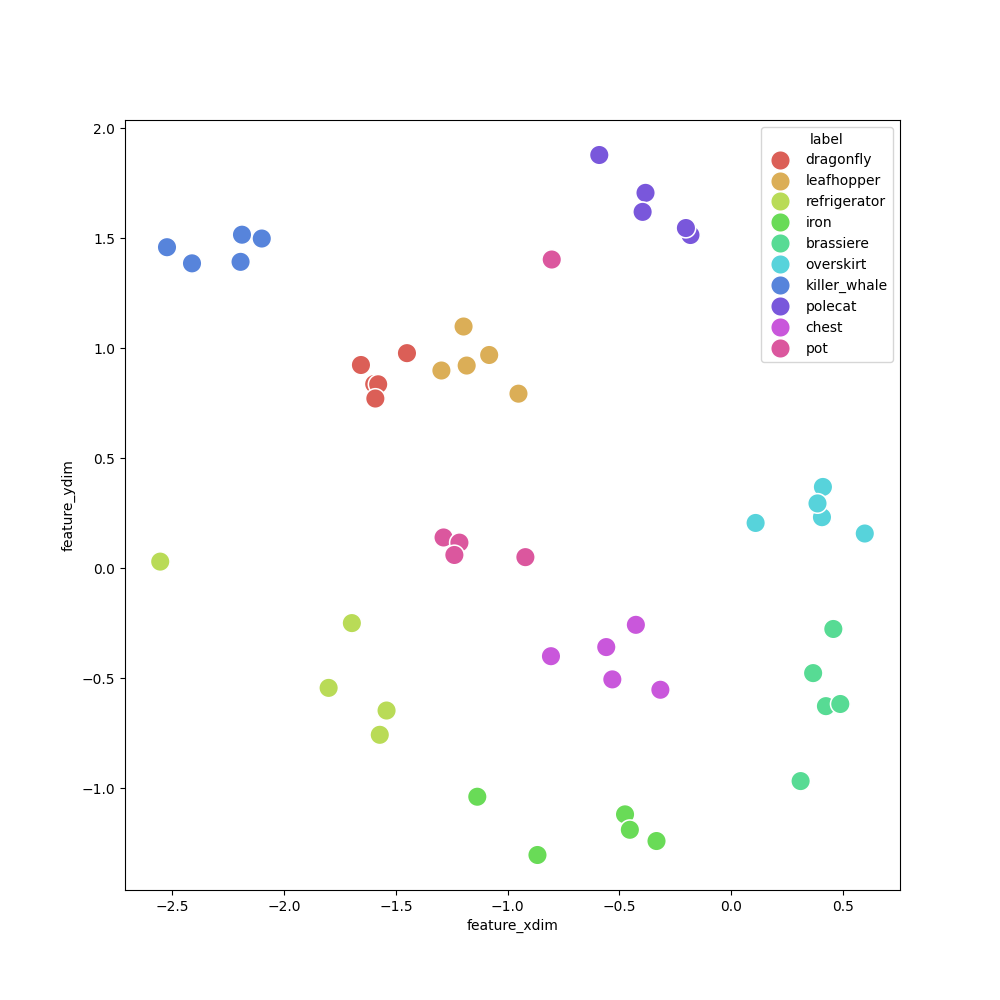}
         \caption{t-SNE scatter plots of feature embeddings from the teacher model ViT-B/16}
         \label{app:fig:final1_img}
\end{subfigure}
\hfill
 \begin{subfigure}{0.49\linewidth}
         \centering
         \includegraphics[width=\textwidth]{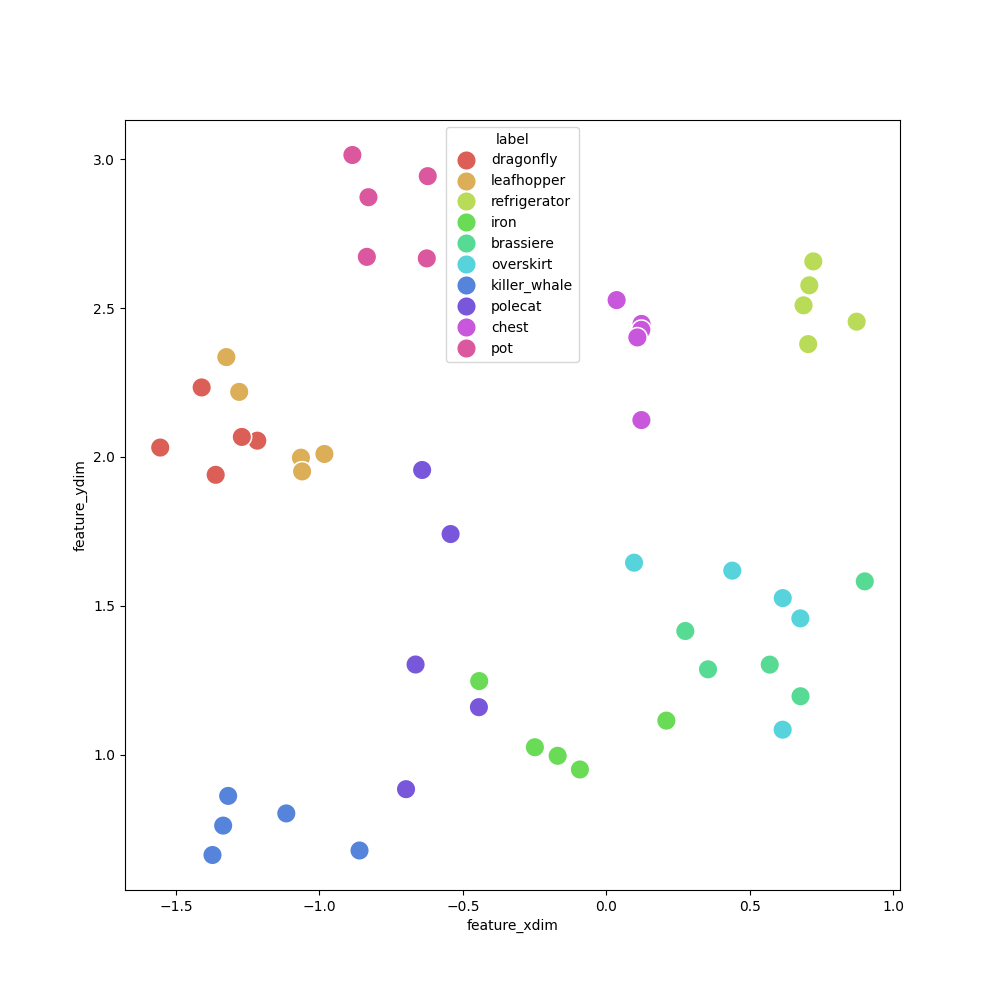}
         \caption{t-SNE scatter plots of feature embeddings from the text encoder with General Prompt}
         \label{app:fig:final1_nogt}
\end{subfigure}
\\
 \begin{subfigure}{0.49\linewidth}
         \centering
         \includegraphics[width=\textwidth]{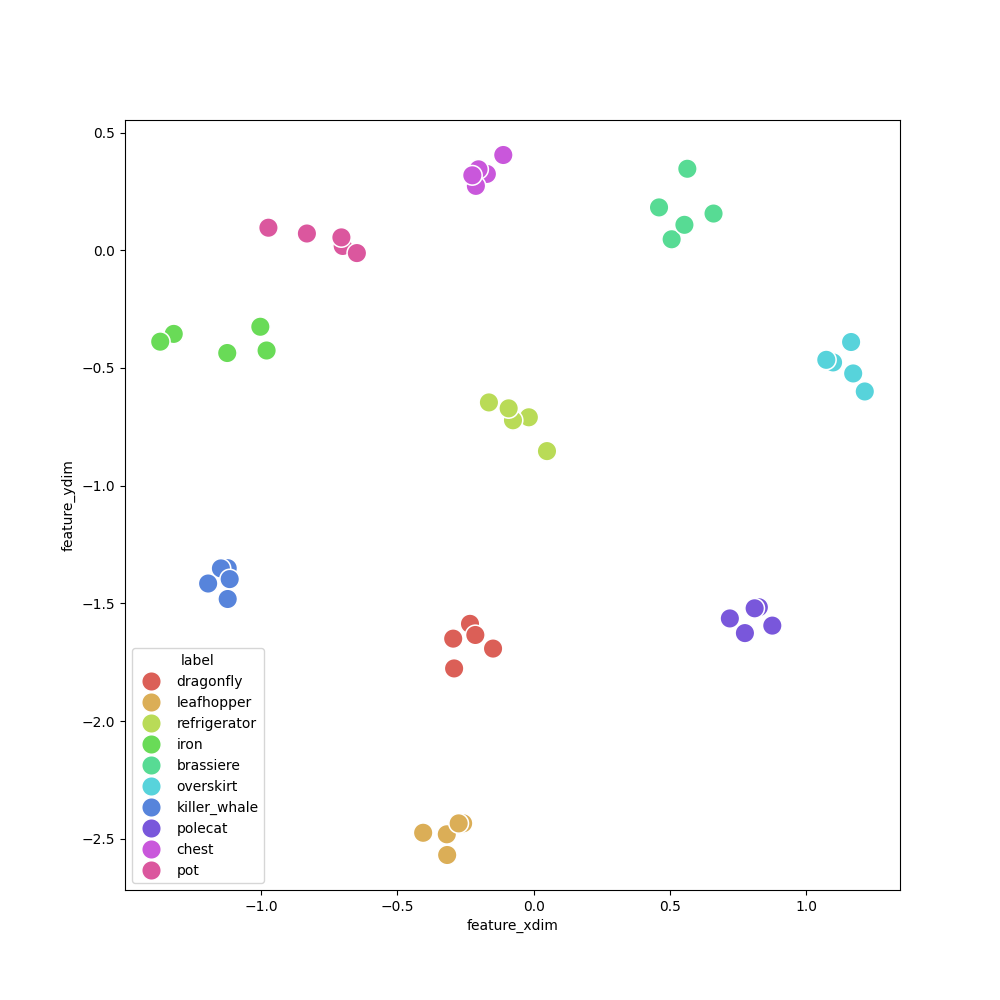}
         \caption{t-SNE scatter plots of feature embeddings from the text encoder with Targeted Prompt (T1)}
         \label{app:fig:final1_gt}
\end{subfigure}
\hfill
 \begin{subfigure}{0.49\linewidth}
         \centering
         \includegraphics[width=\textwidth]{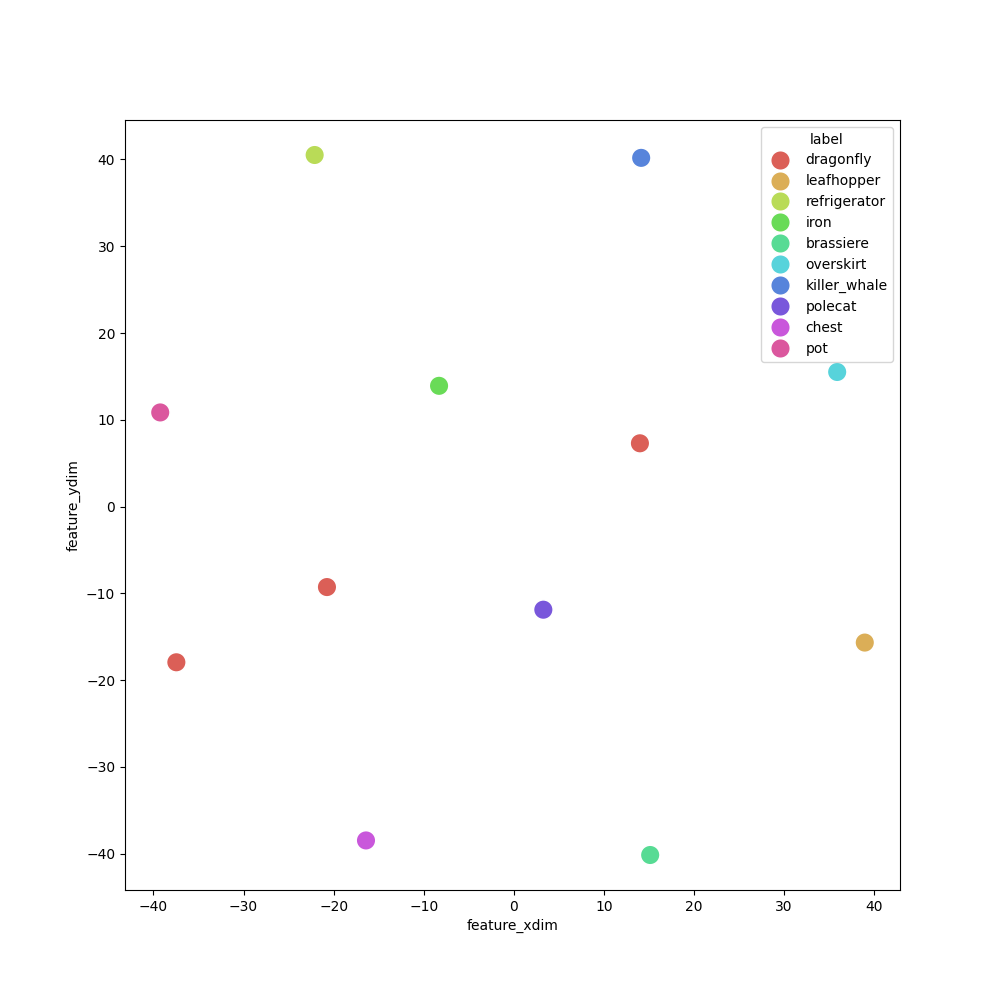}
         \caption{t-SNE scatter plots of feature embeddings from groundtruth classnames}
         \label{app:fig:final1_cls}
\end{subfigure}
\caption{Feature visualization for embeddings across 10 classes. (Best viewed in color.)}
\label{app:img:feat1}
\end{figure*}

 \begin{figure*}[b!]
         \centering
         \includegraphics[width=\textwidth]{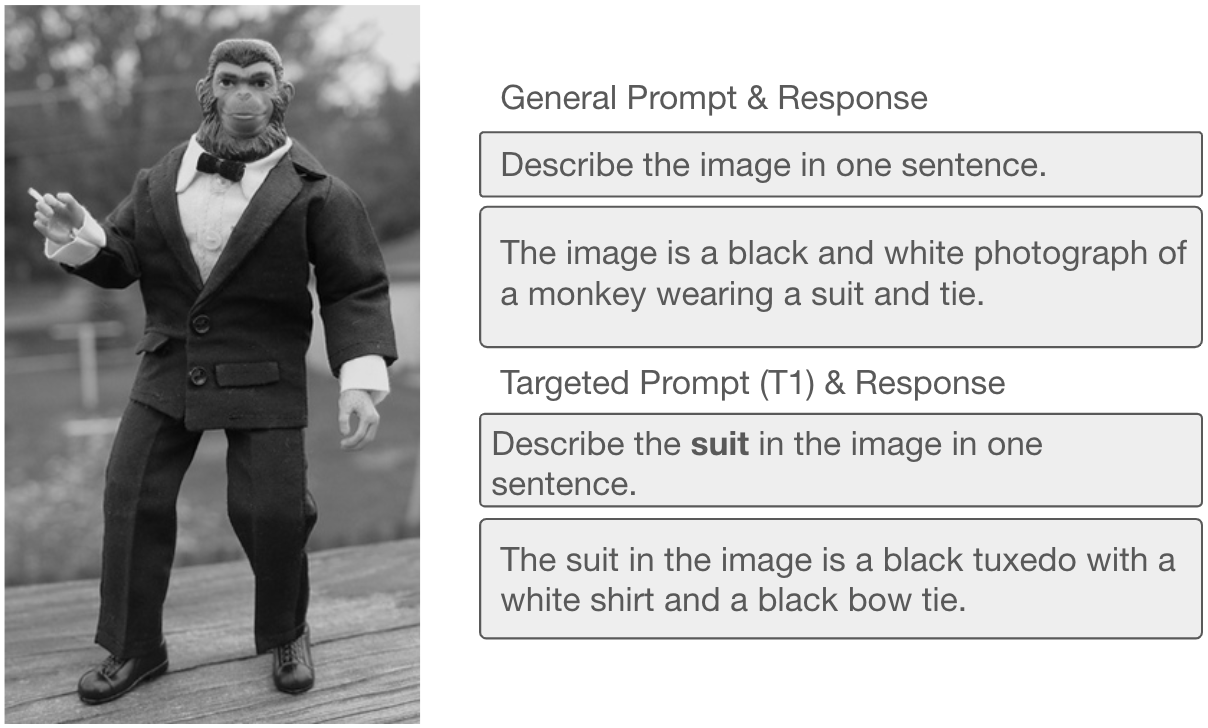}
         \caption{Examples of the prompts used and responses.}
         \label{app:fig:exp_img1}
\end{figure*}

 \begin{figure*}[b!]
         \centering
         \includegraphics[width=\textwidth]{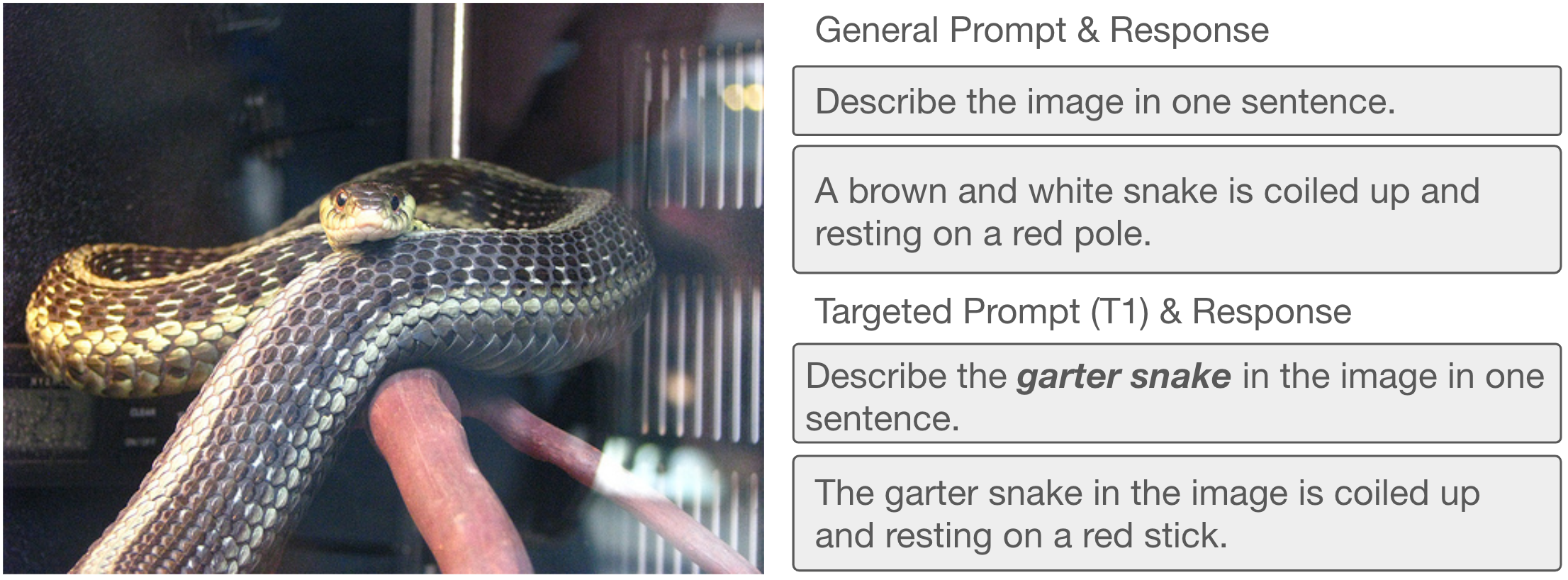}
         \caption{Examples of the prompts used and responses.}
         \label{app:fig:exp_img1}
\end{figure*}

 \begin{figure*}[b]
         \centering
         \includegraphics[width=\textwidth]{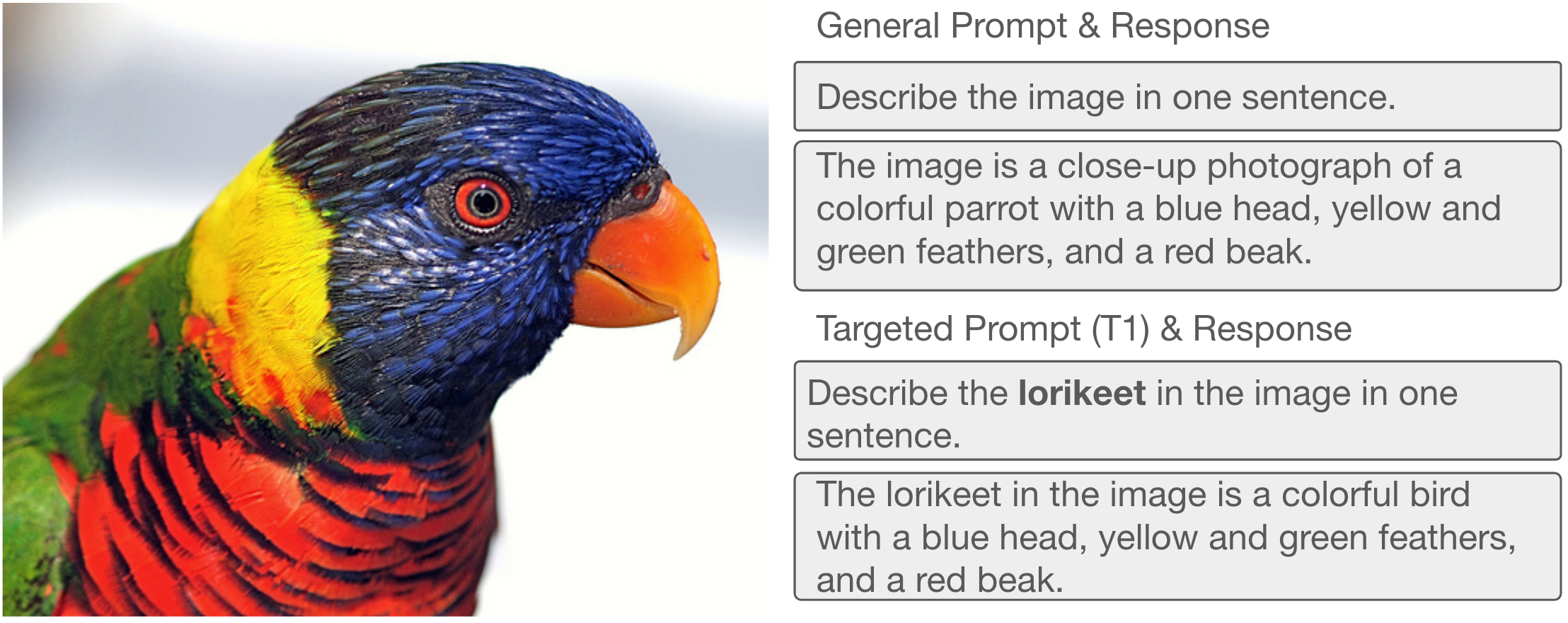}
         \caption{Examples of the prompts used and responses.}
         \label{app:fig:exp_img1}
\end{figure*}

 \begin{figure*}[b]
         \centering
         \includegraphics[width=\textwidth]{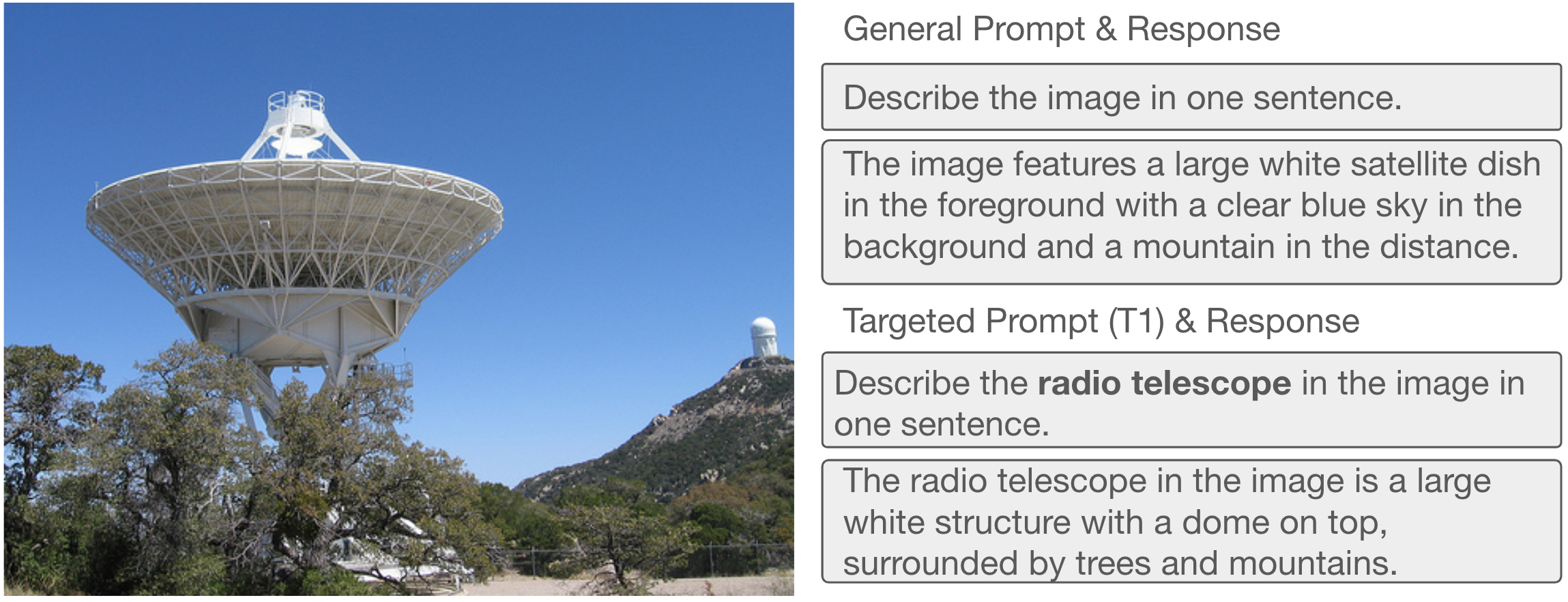}
         \caption{Examples of the prompts used and responses.}
         \label{app:fig:exp_img1}
\end{figure*}

 \begin{figure*}[b]
         \centering
         \includegraphics[width=\textwidth]{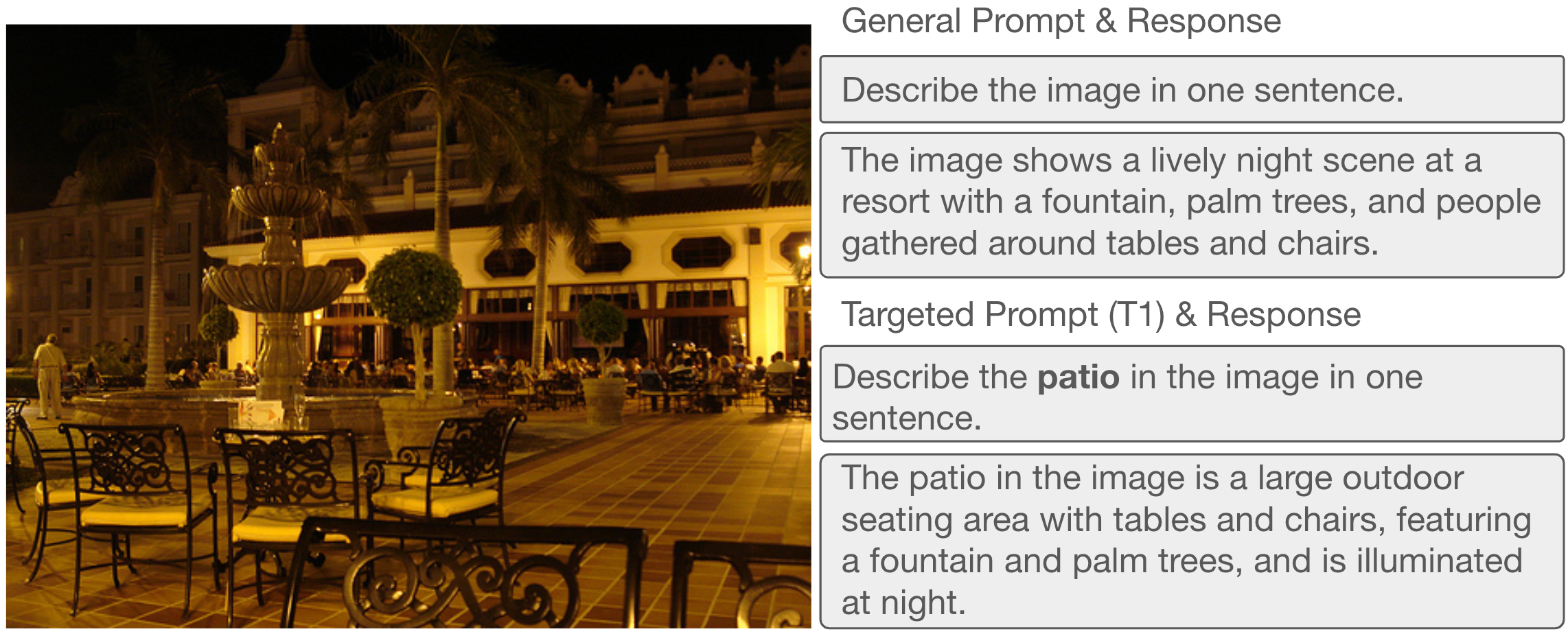}
         \caption{Examples of the prompts used and responses.}
         \label{app:fig:exp_img1}
\end{figure*}

 \begin{figure*}[b]
         \centering
         \includegraphics[width=\textwidth]{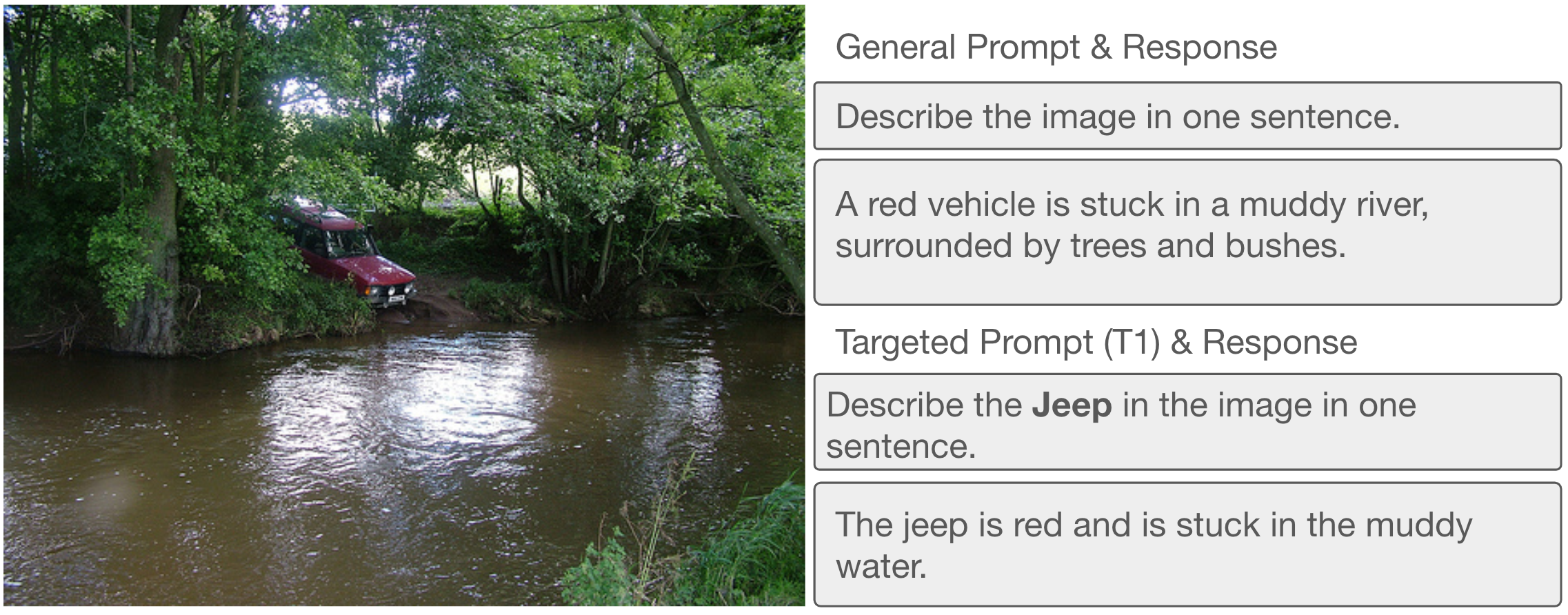}
         \caption{Examples of the prompts used and responses.}
         \label{app:fig:exp_img1}
\end{figure*}

 \begin{figure*}[b]
         \centering
         \includegraphics[width=\textwidth]{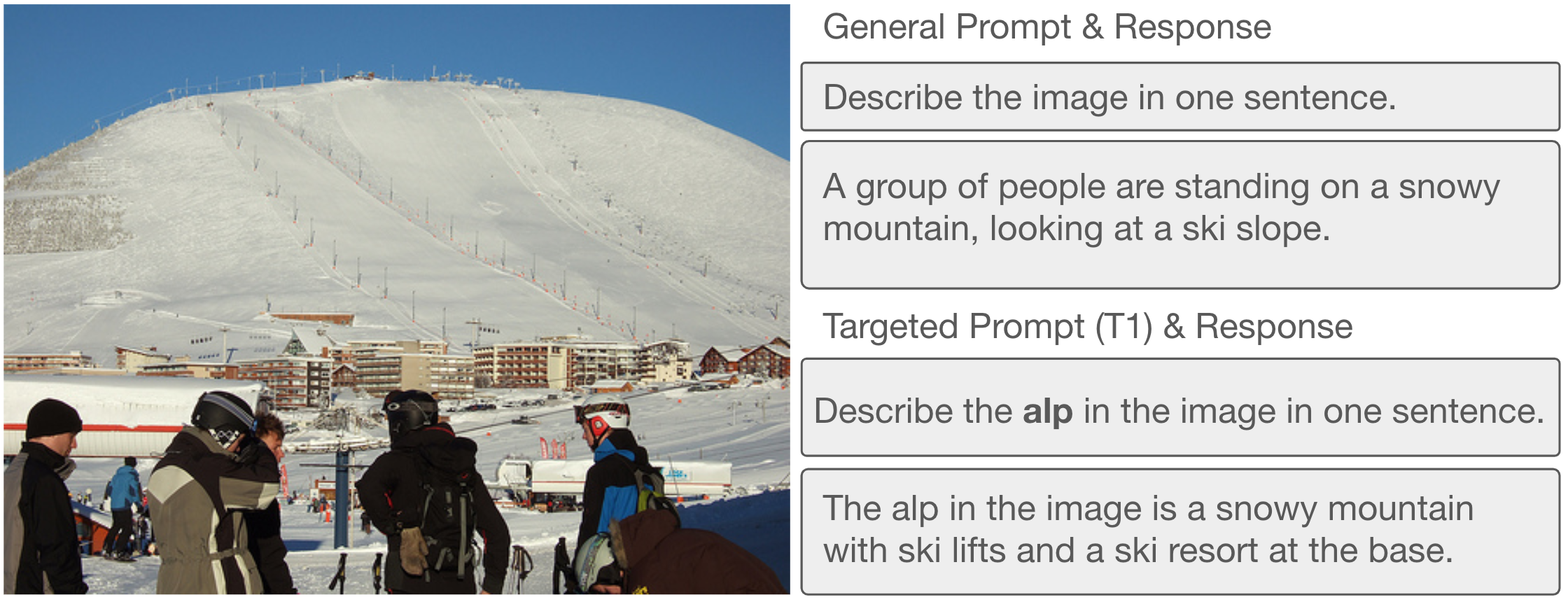}
         \caption{Examples of the prompts used and responses.}
         \label{app:fig:exp_img1}
\end{figure*}

 \begin{figure*}[b]
         \centering
         \includegraphics[width=\textwidth]{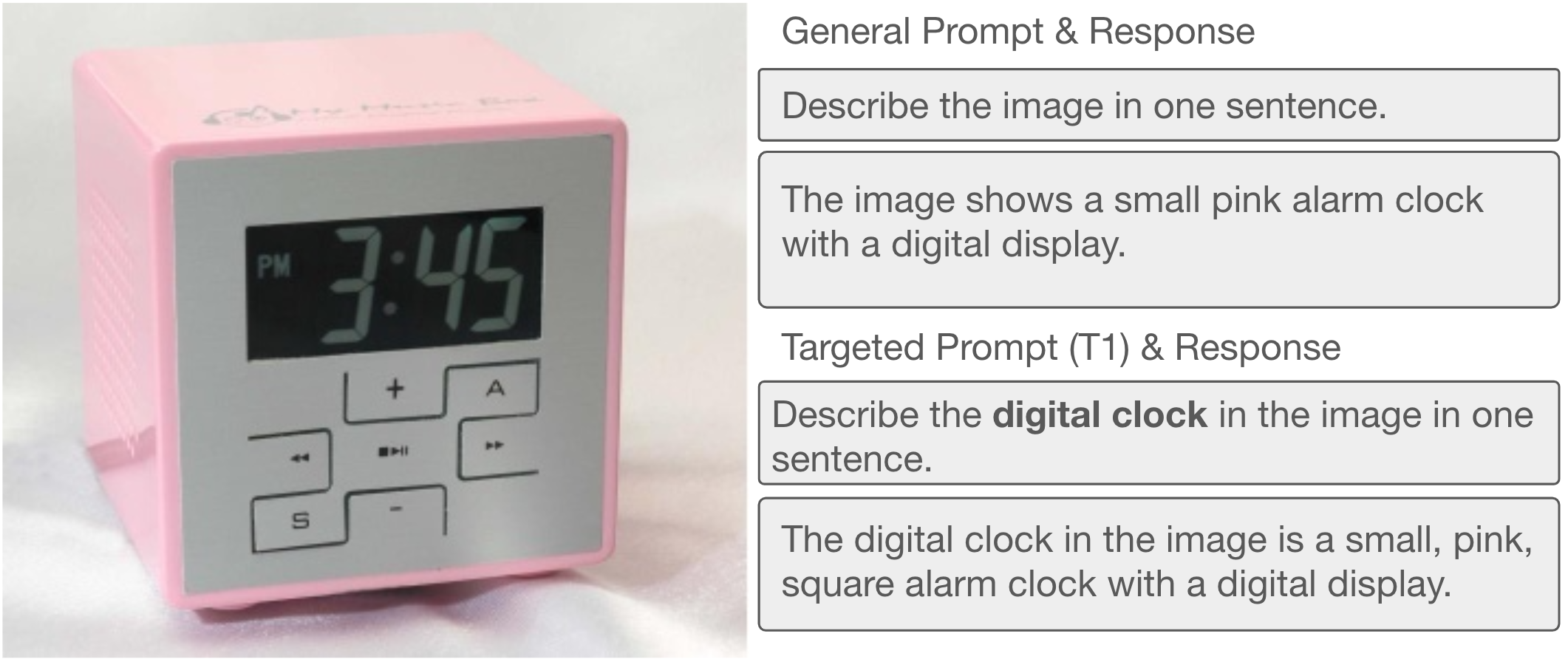}
         \caption{Examples of the prompts used and responses.}
         \label{app:fig:exp_img1}
\end{figure*}

\end{document}

%% file: 01_intro.tex
\section{Introduction}
\label{intro}
Data distributions in real-world contexts often follow a long-tail pattern, marked by an exponential decrease in the number of samples per class from the head to the tail~\cite{van2018inaturalist}. This data imbalance presents a significant challenge for training deep models, as their ability to generalize to infrequent categories can be hindered by the limited training data available for these classes. Since such distributions are inherit in many domains, including wildlife monitoring, medical diagnosis and self-driving scene understanding, addressing the long-tail distribution issue is crucial for deploying deep models in real-world scenarios.  

\input{Figures/lt_vis}

To address the long-tail data problem, several data-driven approaches have been developed. Prior work has implemented class-balanced sampling during training~\cite{wallace2011class}, incorporated weight balancing for long-tail classes~\cite{menon2020long}, and added regularization through data augmentation~\cite{li2021metasaug}. Recent studies have explored promising directions for using knowledge distillation to enhance long-tail understanding, which have focused on training experts with class-balanced data and then distilling their knowledge into a single student model~\cite{xiang2020learning, wang2020long, he2021distilling, li2022nested}. With the recent success of foundational visual feature encoders, researchers are now experimenting with distilling visual knowledge from these foundational models into long-tail classifiers~\cite{huang2023sentence} or directly fine-tuning a pretrained foundational model backbone~\cite{tian2022vl}. However, due to the large datasets used in training foundational models, these models are typically very large and may not be suitable for latency-sensitive applications, such as self-driving vehicles or indoor robots. Additionally, since foundational models are often built with modern architectures like ViTs~\cite{dosovitskiy2020image}, distilling ViT encoder features into convolution-based vision models like ResNet~\cite{he2016deep} can result in reduced benefits due to architectural differences~\cite{wang2023improving}.

In this paper, we explore an alternative supervision signal provided by foundation models: text descriptions. Recent developments in vision-language models (VLMs), such as GPT-4~\cite{achiam2023gpt}, have significantly advanced the ability of AI systems to understand and describe visual content in natural language. By integrating images and prompts, VLMs can effectively describe common object appearances and some detailed visual features. As shown in Figure~\ref{fig:lt_vis}, textual supervision allows us to grasp the compositional nature of long-tail objects, which exhibit common features across multiple categories. 
Moreover, with the availability of open-source VLM models~\cite{liu2024llavanext}, acquiring high-quality text supervisions is becoming increasingly cost-effective.

We propose VLM-KD: a simple yet effective approach to distill knowledge from vision-language models into an image classifier. We begin by demonstrating how to generate text supervision and then how to integrate it into a training procedure. Next, we showcase the effectiveness of our approach across various benchmarks and with different model architectures. We compare our method against several baseline knowledge distillation methods, showing that our approach provides additional benefits with more textual descriptions. Additionally, we conduct a thorough ablation study to assess each component of our method.

\input{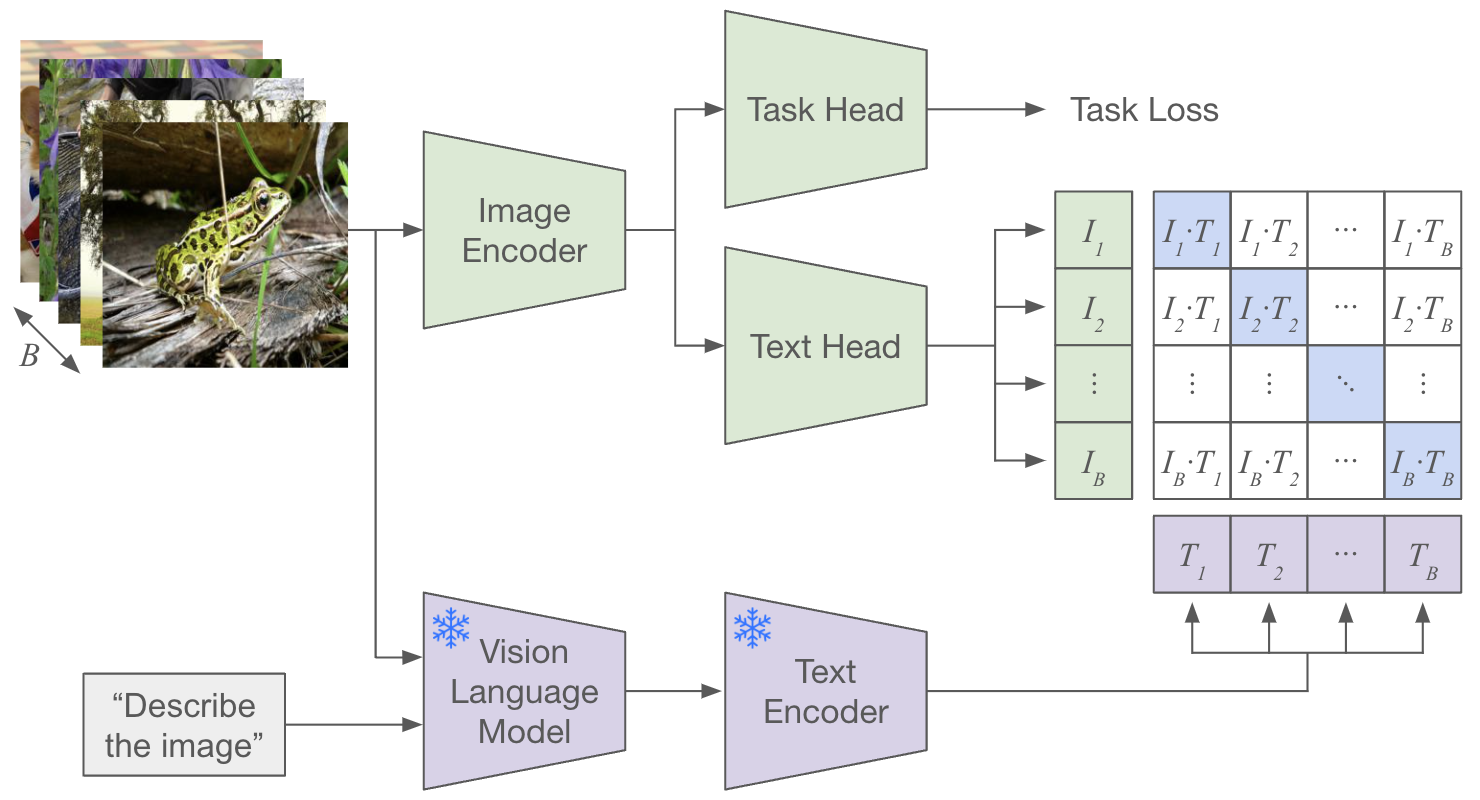}

In this paper, we make the following contributions:
\begin{itemize}
    \item We propose a novel knowledge distillation approach with text supervisions generated by off-the-shelf vision-language models.
    \item We show that knowledge distillation with text features provides significant benefits for different neural network models on multiple long-tail classification benchmarks. 
    \item We demonstrate that features from text supervisions provide complementary benefits in addition to image features during knowledge distillation.
    \item We illustrate that our distillation method scales well with additional text supervisions. 
\end{itemize}

%% file: Figures/lt_vis.tex
\begin{figure}[t]
     \centering
     \includegraphics[width=0.95\columnwidth]{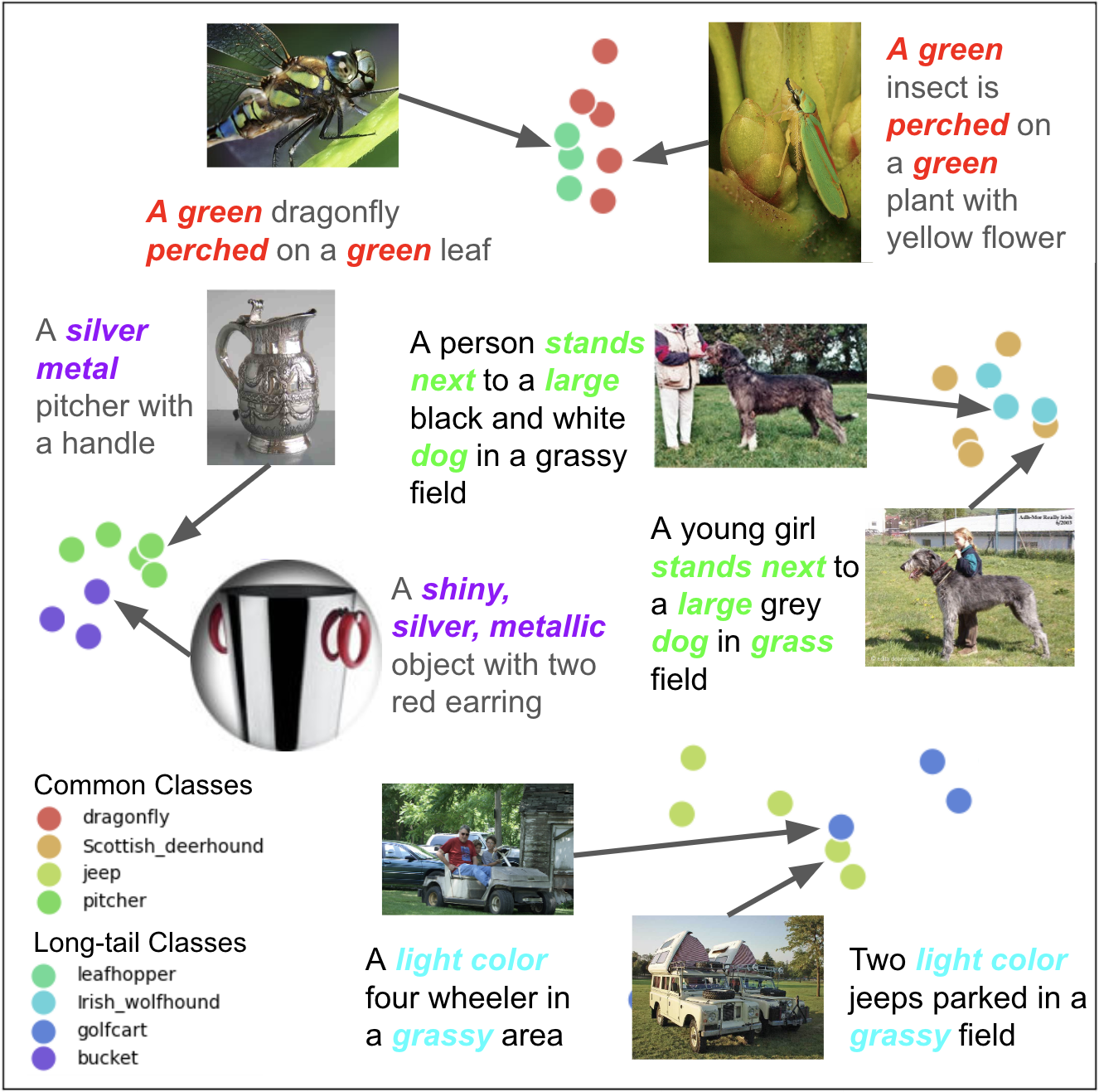}
     \caption{t-SNE scatter plots of feature embeddings from sampled text generated by a VLM. Instances from long-tail classes are joining feature clusters formed by common categories. Grouped features exhibit strong semantic relevance.}
     \label{fig:lt_vis}
\end{figure}

%% file: Figures/overview.tex
\begin{figure*}[t]
     \centering
     \includegraphics[width=\textwidth]{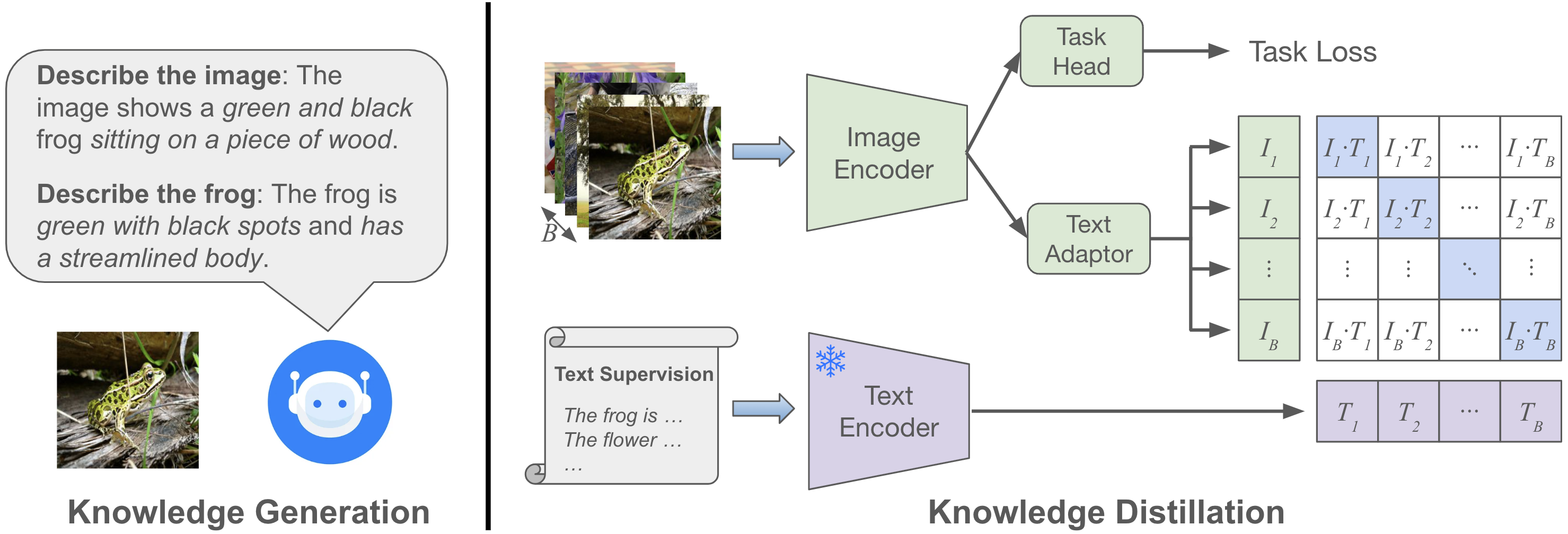}
     \caption{Overview of VLM-KD. We first query VLMs with images and prompts to generate text supervisions for each image in the dataset. Then we encode the text with a pretrained text encoder. Note that text feature supervisions are pre-computed for the entire dataset before training. Finally, with paired image and text features, we add a constrastive loss in the original classifier training to distill text semantics into the image encoder. The text adaptor is discarded after training.}
     \label{fig:overview}
\end{figure*}

%% file: 02_related.tex
\section{Related work}
\label{related}

\paragraph{Long-tail Visual Recognition} Early rebalancing strategies for the long-tail recognition mainly fall into two main types: resampling and reweighting. Resampling helps mitigate the skewed distribution of training data by either undersampling~\cite{kubat1997addressing,wallace2011class} or oversampling~\cite{chawla2002smote}. In contrast, reweighting adjusts the loss function to enhance the influence of long-tail classes during the training process~\cite{huang2016learning, menon2013statistical}. Additionally, two prevalent approaches have emerged: post-hoc normalization of classifier weights and modification of the loss margin. The motivation behind post-hoc normalization is that the norm of classifier weights often mirrors the class distribution, and this can be rectified by normalizing these weights~\cite{kang2019decoupling,kim2020adjusting,zhang2019balance}. On the other hand, loss margin modification~\cite{cao2019learning,menon2020long,ren2020balanced,tan2020equalization} introduces prior knowledge of class distribution by adjusting the classifier’s margin, with techniques like Logit Adjustment~\cite{menon2020long} and Balanced Softmax~\cite{ren2020balanced} effectively recalibrating the decision boundary based on class probabilities. Other widespread methods~\cite{he2016deep,wang2021regularizing,li2021metasaug,chu2020feature,zang2021fasa,wang2021revisiting,wang2023efficienttrain,huang2022glance} involve enhancing the representation of minority classes through data augmentation.

Recently, researchers have utilized contrastive learning to address the challenges of long-tail recognition. Contrastive learning, a self-supervised learning technique, employs a contrastive loss function to enhance data representations by maximizing the similarity between contrasting (positive and negative) samples~\cite{chen2020simple,grill2020bootstrap,tian2020contrastive,he2020momentum}. Khosla~\cite{khosla2020supervised} \emph{et al.} expanded this concept into the supervised contrastive learning (SCL) framework by including label information, thereby aligning learning more closely with supervised tasks.
However, contrastive learning can be compromised by the imbalance between positive and negative samples, often causing models to disproportionately focus on the more prevalent head categories in long-tail datasets~\cite{fang2021exploring,cui2021parametric,kang2020exploring}. To mitigate this, several studies~\cite{li2022targeted,zhu2022balanced,cui2021parametric,wang2021contrastive,samuel2021distributional,cui2023generalized,du2024probabilistic} have suggested the use of class complement strategies for forming these pairs. These methods ensure that all classes are equally represented in every training iteration, thus helping to evenly distribute the contrast samples across different categories.

\paragraph{Knowledge Distillation for Visual Recognition} Knowledge distillation is a technique in which a student model is trained by mimicking the outputs of a well trained teacher model~\cite{hinton2015distilling}. Work in this area can be categorized into two main streams. The first stream emphasizes the design of loss functions applied to the outputs of both the teacher and student models, employing losses like $l_1$~\cite{kim2018paraphrasing}, $l_2$~\cite{chen2020learning,passban2021alp,wang2020exclusivity}, Maximum Mean Discrepancy (MMD)~\cite{huang2017like}, Kullback-Leibler (KL) divergence~\cite{chen2018darkrank,passalis2020heterogeneous,passalis2020probabilistic}, and cross-entropy losses~\cite{liu2019knowledge,xu2020feature}. The second stream concentrates on aligning the intermediate feature representations from specific layers~\cite{guan2020differentiable,heo2019comprehensive,shen2019amalgamating,yang2020knowledge}, or functions derived from these embeddings~\cite{li2021micro,yim2017gift,zhang2018better}, to facilitate deeper similarities beyond the final output.

Recently, several techniques have been specifically designed to conduct knowledge distillation techniques in long-tail learning scenarios. They usually conduct training on multiple experts that focus on different, more balanced data or subsets of samples (like head, middle, and tail groups). Then distill the insights from these experts into a singular, comprehensive student model~\cite{xiang2020learning, wang2020long, he2021distilling, li2022nested}. In this paper, we focus on knowledge distillation from foundation models, which are trained with large and diverse datasets.

\paragraph{Vision-Language Model}
Vision-language representation learning has shown significant effectiveness across various downstream vision tasks~\cite{lu2019vilbert, tan2019lxmert, chen2019uniter, li2020closer, li2020oscar, zhang2021vinvl,li2020hero,gan2020large,lu202012,radford2021learning}. More recently, models like CLIP~\cite{radford2021learning} and ALIGN~\cite{jia2021scaling} have advanced visual-language representation learning through contrastive learning techniques applied to large-scale image-text pairs.

Motivated by these successes, researchers are now applying multi-modal foundational models to address long-tail recognition challenges. For instance, VL-LTR~\cite{tian2022vl} has created a class-level visual-language pre-training method that links images with textual descriptions at the class level. It also integrates a language-guided recognition head, which effectively utilizes visual-linguistic representations to improve visual recognition. BALLAD~\cite{ma2021simple} also conducts visual-language pre-training and then finetunes an adaptor on a class-balanced dataset to improve the long-tail visual recognition. In contrast to the methods mentioned above that require a pre-trained CLIP image encoder, Huang~\cite{huang2023sentence} \emph{et al.} concentrate on transferring the knowledge from both CLIP text and vision encoders to a student model that is initialized randomly. Similarly, we also focus on knowledge distillation to any vanilla student model. While previous methods rely solely on either ground truth class names~\cite{huang2023sentence} or a snippet from Wikipedia for each class~\cite{tian2022vl}, our approach demonstrates the advantages of distilling knowledge from descriptions generated by a VLM for each image.

%% file: 03_approach.tex
\section{Approach}
\label{app}

In this section, we present the details of our approach for distilling knowledge from an off-the-shelf Vision-Language Model (VLM) into a classifier model. An overview of our approach is shown in Figure~\ref{fig:overview}. We first state the problem setup and training objectives for the long-tail image classification task, then we describe how to generate the text knowledge, and finally show how the model can learn from the generated text.

\subsection{Preliminaries}
Given a dataset $D$, with a collection of $N$ images $I$ and their corresponding label set $Y$, the objective of an image classifier is to associate an image ${x \in I}$ to a label ${y \in Y}$ with a mapping function. Typically, the mapping function is
modeled as a deep neural network, which consists of a backbone feature extractor $\theta$ and a linear classifier $H$. For a given image $x$, prediction ${Z = H(\theta(x))}$ represents the un-normalized class distribution over all possible classes. We use the Logit Adjustment~\cite{menon2020long} loss function to supervise the classification training objective. Logit Adjustment uses the prior probability of each class as a weight throughout the training and inference stages. Given an image $x$ and corresponding label $y$, the loss is defined as:
\begin{equation}
    L_{cls} = - \log\frac{\pi_y\exp({H_y(\theta(x))})}{\sum\limits_{y' \in Z}\pi_{y'}\exp({H_{y'}(\theta(x))})}
\label{equ:cls}
\end{equation}
where ${\pi_y}$ is the class frequency in the training set and $H_y(\theta)$ is the predicted logit of the class $y$.

Supervised Contrastive Loss has been widely applied to many modern classification models~\cite{cui2021parametric,cui2023generalized,li2022targeted,du2024probabilistic}. It is typically designed to train a feature extractor $\theta$ that differentiates between positive pairs from the same class and negative pairs from different classes. In this paper, we also deploy this loss during training and denote it as ${L_{scl}}$. We adopt the loss formulations from ~\cite{cui2023generalized,du2024probabilistic} and apply accordingly based on the target benchmark. More details are in Appendix A.1.

\subsection{Text supervision generation}
Recent advancements in Large Vision-Language Models (VLMs) like GPT-4~\cite{achiam2023gpt}, Gemini-Pro~\cite{reid2024gemini}, and others~\cite{liu2024llavanext, chen2023sharegpt4v} have made it feasible to efficiently generate text supervisions with varying levels of semantic understanding on a large scale. For our text supervision generation, we chose the open-source VLM model, LLaVA-NeXT~\cite{liu2024llavanext}. To minimize computational costs, we opted for the version of LLaVA-NeXT that utilizes the Mistral-7B decoder~\cite{jiang2023mistral}. We have observed that in terms of distilling knowledge into an image classifier, larger VLMs do not provide substantial advantages over smaller models, as demonstrated in Section~\ref{abla:vlm}. For a given dataset $D$ with $N$ images, we denote the generated text datasets as ${[D_t^1, D_t^2, ...D_t^Q]}$, where in each $D_t^i$, there are $N$ corresponding text descriptions. $Q$ represents the number of input queries.
\paragraph{Prompt Tuning for Dataset Generation} To extract textual knowledge from an image, we mainly use two types of prompts: ($1$) General Prompt: describe the various visual features from the image; ($2$) Targeted Prompt: describe the targeted object in the image by injecting groundtruth into the prompts. Example prompts and responses are shown in Figure~\ref{fig:overview}. More examples can be found in Appendix A.4. We find these two types of prompt provide complementary benefits, see more details in Section~\ref{exp:scale}.

\subsection{Knowledge distillation through text} 


Drawing inspiration from the success of CLIP~\cite{radford2021learning}, we propose adding a contrastive learning objective during training. This objective aims to align the image embedding with the embedding of its corresponding text description, while pushing it away from the embeddings of other text descriptions within the same batch. Given a batch of $B$ images with their corresponding text descriptions ${[(x_1, T_1), . . .(x_B, T_B)]}$, the new objective is defined as follows:
\begin{equation}
    L_{text} = - \sum_{i=1}^B\log\frac{\exp(\bar{F}(x_i)\cdot\bar{G}(T_i))/\tau}{\sum_{j=1}^B \exp(\bar{F}(x_i)\cdot\bar{G}(T_j))/\tau}
\label{equ:text}
\end{equation}
where ${F(x) = A(\theta(x))}$ and $A$ is a text adaptor to project the image embeddings closer to the features in the text embedding space. $G$ is an off-the-shelf text encoder to encode the free-form text descriptions into a feature embedding. In practice, ${G(T)}$ is pre-computed and cached. ${\bar{F}}$ and ${\bar{G}}$ indicate that the output of $F$ and $G$ are ${l2}$ normalized. Like in ~\cite{wu2018unsupervised}, $\tau$ is a temperature parameter.

\noindent\textbf{Text Encoder} We use the pre-trained CLIP text encoder ViT-B/16 to pre-compute the feature embeddings for $G$. However, as discussed in Section~\ref{abla:text}, our method is flexible and does not depend on a specific text encoder.

\noindent\textbf{Text Adaptor} We apply a learnable nonlinear transformation on the image embeddings prior to aligning them with text features. This transformation consists of a linear layer, batch normalization, and ReLU activation, followed by another linear layer. We demonstrate the importance of this adaptor in Section~\ref{abla:ada}.

\subsection{Loss formulation} 
Given the original dataset $D$ and generated text datasets ${[D_t^1, D_t^2, ...D_t^Q]}$, our final loss is defined as:
\begin{equation}
    L = L_{cls} + L_{scl} + \alpha \sum_{k=1}^{Q}L_{text}.
\label{equ:text}
\end{equation}
In our experiments, we use a learnable temperature parameter $\tau$, which is initialized to 0.07, and we set $\alpha$ to 1 for all our experiments.

%% file: 04_result.tex
\section{Experimental results}
\label{exp}
In this section, we assess the effectiveness of our approach across various benchmarks and settings, comparing it with existing knowledge distillation methods. First, we detail the benchmarks used, followed by the implementation details for both the baseline methods and our approach. Next, we present the effectiveness of our approach across all benchmark datasets and different neural network models. Finally, we illustrate the scaling effects of text supervision.

\input{Tables/main_imglt}

\subsection{Dataset and Evaluation}
\label{exp:data}
We conduct long-tail image classification experiments on three widely used datasets: ImageNet-LT~\cite{liu2019large}, iNaturalist 2018~\cite{van2018inaturalist} and Place-LT~\cite{liu2019large}. Following~\cite{liu2019large}, we divide all categories into three subsets based on the number of training samples: Many ($>$ 100 images), Medium (20 to 100 images), and Few ($<$ 20 images). We report the top-1 accuracy on the corresponding balanced validation sets.
\paragraph{ImageNet-LT} ImageNet-LT~\cite{liu2019large} is created by sub-sampling ImageNet~\cite{ILSVRC15} based on a Pareto distribution with a power value of 6. It comprises of 115.8k images across 1,000 categories, with class cardinality ranging from 5 to 1,280.
\paragraph{iNaturalist 2018} iNaturalist 2018~\cite{van2018inaturalist} is a highly imbalanced dataset, containing 437.5k images spanning 8,142 classes with class cardinality ranging from 2 to 1,000. 
\paragraph{Places-LT} Places-LT is a long-tail variant of the large-scale scene classification dataset Places~\cite{zhou2017places}. It contains 184.5k images across 365 categories, with class cardinality ranging from 5 to 4,980.

\subsection{Baseline Methods}
\label{exp:base}
\paragraph{KD-Image} KD-Image (referred to as KD-I) combines both logit distillation~\cite{hinton2015distilling} and feature distillation~\cite{huang2023sentence} to form a strong baseline with a vision-only teacher model. For fair comparison, we finetune a pre-trained CLIP vision encoder ViT-B/16 to produce a vision-only teacher classifier model. Given a student classifier $S$ and a teacher classifier $T$, we denote the final prediction logits as ${q^S}$ and ${q^T}$. For feature distillation, due to the major architecture differences in our setting, we only use the last hidden layer embeddings for distillation, which are denoted as ${f^S}$ and ${f^T}$. The KD-Image loss is defined as:
\begin{equation}
    L_{kd} = \frac{1}{B}\sum_{i=1}^B {KL}(q^S_i, q^T_i) + \frac{1}{B}\sum_{i=1}^B (f^S_i -  f^T_i)^2
\label{equ:cls}
\end{equation}
where $KL$ is the KL divergence loss and $B$ is the batch size. 

\paragraph{RISE} RISE~\cite{huang2023sentence} distills knowledge to a student model from pre-trained CLIP vision and text encoders. However, they only pass text based on the groundtruth label to the text encoder, such as “a photo of a dog”, etc. For fair comparison, we use the pre-trained CLIP ViT-B/16 as the teacher, and we follow the same hyperparameter search protocols in~\cite{huang2023sentence}.

\input{Tables/main_placelt}

\input{Tables/main_diffarch}

\input{Tables/main_scaling_img}
\subsection{Implementation Details}
\label{exp:imple} 
We set a fixed random seed, conduct all experiments three times and record the mean measurement. We conduct all the experiments on Nvidia H100 GPUs. Additional details can be found in Appendix A.1.

\noindent\textbf{ImageNet-LT} We adopt the training procedure from ProCo~\cite{du2024probabilistic} for ImageNet-LT. We increased the batch size to 1024 and change the initial learning rate to 0.4. We keep other hyper-parameters the same and train the model for 180 epochs. On ImageNet-LT, we show results for multiple backbones, including ResNet-18, ResNet-34, ResNet-50 and ViT-Base-16. For the ViT model, we use a pre-trained model provided by LiVT~\cite{LiVT} and adopt the same configurations for finetuning. 

\noindent\textbf{iNaturalist 2018} For iNaturalist 2018, we adopt the same configurations as specified in ProCo~\cite{du2024probabilistic}, except we set the batch size to 512 and change the initial learning rate to 0.1. We train the model for 90 epochs. 
We conduct experiments with ResNet-50~\cite{he2016deep}.

\noindent\textbf{Places-LT} We adopt the same configurations from GPaCo~\cite{cui2023generalized}, but we increase the batch size to 256 and train the model for 60 epochs. Following the common practice~\cite{cui2023generalized}, we conduct the experiments with ResNet-152 pre-trained on ImageNet-1K~\cite{ILSVRC15}. 

\subsection{Results with General Captions}
\label{exp:main}
ResNet-50~\cite{he2016deep} serves as the main backbone for comparison. The text supervision is generated with the General Prompt: ``\textit{Please describe the image in one sentence.}''

\noindent\textbf{ImageNet-LT} As shown in Table~\ref{table:main_table_imglt}, using text supervision (KD-T) results in a 2.9\% improvement in top-1 accuracy, compared with ProCo~\cite{du2024probabilistic}. Utilizing a fine-tuned CLIP ViT-B/16 as the vision-only teacher model (KD-I) also enhances performance compared to ProCo. However, the best overall results are achieved by combining both image-based and our text-based distillation (KD-I-T). This approach delivers a 3.5\% improvement over the top-performing state-of-the-art methods and a 1.1\% improvement over prior knowledge distillation method with VLMs (RISE).

\noindent\textbf{iNaturalist 2018} For iNaturalist 2018, given the close resemblance among different classes, we modify the prompt slightly to ``\textit{Please describe the image in detail.}'' to extract more textural features. Our method achieves a 0.9\% improvements over our reproduced ProCo~\cite{du2024probabilistic} baseline as shown in Table~\ref{table:main_table_imglt}. 
Combining the two distillation methods still provides additional improvements. This emphasizes that text supervisions provide complementary benefits in addition to image features. 

\noindent\textbf{Places-LT} Our distillation method continues to offer improvements over the reproduced baseline, as shown in Table~\ref{table:main_table_placelt}. 
Our method offers 1.1\% performance improvements over the prior SOTA. This shows that our approach is robust with smaller training dataset.


\subsection{Performance on Different Models}
\label{exp:diff} In Table~\ref{table:main_table_diffarch}, we show the performance of our approach across different network architectures. ResNet models are trained with the ProCo~\cite{du2024probabilistic} settings, and ViT models are trained with the LiVT~\cite{LiVT} settings. Our method provides significant improvements across all architectures: 3.3\% on ResNet-18 and ResNet-34, and 4.9\% on ViT-B/16.  Our method improves performance significantly on the long-tail cases for ResNets (the ``few'' category): 4.9\% for ResNet-18 and 5.3\% for ResNet-34. In fact, our ResNet-34 out-performs all ResNet-50 SOTA approaches on the long-tail categories. This demonstrates the importance of applying our approach to long-tail classification.

\input{Tables/ablade_text}
\input{Tables/ablade_naive}

\subsection{Text Supervision Scaling}
\label{exp:scale}
With various prompts, we can create multiple text datasets. In Figure~\ref{table:main_table_scale}, we explore different methods of incorporating additional text supervision during distillation for two architectures: ResNet and ViT. We use the two prompt styles illustrated in Figure~\ref{fig:overview}. The term ``Concat'' refers to extracting features from a simple concatenation of the text. ``Seperate'' and ``Shared'' involves applying the contrastive loss to each text features and then aggregating the losses, as described in Eq.~\eqref{equ:text}. ``Seperate'' denotes using separate text adaptors for each prompt, while ``Shared'' means using the same text adaptor for all prompts. Our findings suggest that concatenating text descriptions does not enhance performance, as it may dilute distinct features. Additionally, employing multiple text adaptors could lead to overfitting. Based on our observations, applying the contrastive loss to each text prompt using a shared text adaptor yields the best performance. On ViT, the improvements scales well with adding more text supervisions. Refer to Appendix A.2 for more results.



%% file: Tables/main_imglt.tex
\begin{table*}[t]
  \centering
  \begin{tabular}{p{4.75cm}p{1.0cm}p{1.0cm}p{1.0cm}p{1.1cm}|p{1.0cm}p{1.0cm}p{1.0cm}p{1.0cm}}
    \toprule
    & \multicolumn{4}{c}{ImageNet-LT} & \multicolumn{4}{c}{iNaturalist 2018} \\
    \cmidrule(r){2-9}
    Method & Many & Med & Few & Avg & Many & Med & Few & Avg \\
    \midrule
    $\tau$ -norm~\cite{kang2019decoupling} & 56.6 & 44.2 & 27.4 & 46.7 & 65.6 & 65.3 & 65.9 & 65.6  \\
    DisAlign~\cite{zhang2021distribution} & 61.3 & 52.2 & 31.4 & 52.9 & 69.0 & 71.1 & 70.2 & 70.6 \\
    GCL~\cite{li2022long} & 63.0 & 52.7 & 37.1 & 54.5 & 67.5 & 71.3 & 71.5 & 71.0 \\
    RIDE~\cite{wang2020long} & 66.2 & 51.7 & 34.9 & 54.9 & 72.2 & 70.2 & 72.2 & 72.7 \\
    BCL~\cite{zhu2022balanced} & - & - & - & 56.0 & 66.7 & 71.0 & 70.7 & 70.4 \\
    DOC~\cite{wang2022towards} & 65.1 & 52.8 & 34.2 & 55.0 & 72.8 & 71.7 & 70.0 & 71.0 \\
    DLSA~\cite{xu2022constructing} & 67.8 & 54.5 & 38.8 & 57.5 & - & - & - & 72.8 \\
    PaCo~\cite{cui2021parametric} & 64.4 & 55.7 & 33.7 & 56.0 & 70.3 & 73.2 & 73.6 & 73.2 \\
    NCL~\cite{li2022nested} & 67.3 & 55.4 & 39.0 & 57.7 & 72.0 & 74.9 & 73.8 & 74.2 \\
    GPaCo~\cite{cui2023generalized} & - & - & - & 58.5 & 73.0 & 75.5 & 75.7 & 75.4 \\
    ProCo~\cite{du2024probabilistic} & 68.2 & 55.1 & 38.1 & 57.8 & \textbf{74.0} & \textbf{76.0} & \textbf{76.0} & \textbf{75.8} \\
    \midrule
    ProCo* & 67.8 & 55.1 & 39.2 & 57.8 & 70.8 & 73.2 & 72.6 & 72.7 \\
    ProCo* + RISE & 70.8 & 58.4 & 41.7 & 60.9 & 70.8 & 74.3 & 73.0 & 73.4 \\
    ProCo* + KD-T & 70.6 & 58.0 & 42.1 & 60.7 & 70.8 & 74.4 & 73.3 & 73.6 \\
    ProCo* + KD-I & 70.7 & 58.3 & 41.6 & 60.8 & 70.9 & 74.5 & 73.2 & 73.6\\
    ProCo* + KD-I-T & \textbf{71.0} & \textbf{60.1} & \textbf{43.2} & \textbf{62.0} & 71.0 & 74.5 & 74.2 & 74.0\\
    \midrule
    Vision-only teacher model & \textcolor{gray}{82.2} & \textcolor{gray}{73.9} & \textcolor{gray}{52.8} & \textcolor{gray}{74.4} & \textcolor{gray}{78.0} & \textcolor{gray}{75.1} & \textcolor{gray}{71.6} & \textcolor{gray}{74.0} \\
    \bottomrule
  \end{tabular}
  \caption{Comparisons on ImageNet-LT and iNaturalist 2018 trained with ResNet-50. (* indicates our reproduced result.)}
  \label{table:main_table_imglt}
\end{table*}

%% file: Tables/main_placelt.tex
\begin{table}[b!]
  \centering
  \begin{tabular}{p{4cm}p{0.5cm}p{0.5cm}p{0.5cm}p{0.5cm}}
    \toprule
    Method & Many & Med & Few & Avg \\
    \midrule
    $\tau$ -norm~\cite{kang2019decoupling} & 37.8 & 40.7 & 31.8 & 37.9 \\
    Bal-Softmax & \textbf{42.0} & 39.3 & 30.5 & 38.6 \\
    ResLT~\cite{cui2022reslt} & 39.8 & 43.6 & 31.4 & 39.8 \\
    MiSLAS~\cite{zhong2021improving} & 39.6 & 43.3 & 36.1 & 40.4 \\
    PaCo~\cite{cui2021parametric} & 36.1 & 47.9 & 35.3 & 41.2 \\
    GPaCo~\cite{cui2023generalized} & 39.5 & 47.2 & 33.0 & 41.7 \\
    \midrule
    GPaco* & 36.1 & 48.0 & 36.5 & 41.5 \\
    GPaco* + RISE & 36.6 & 48.5 & 37.0 & 42.0 \\
    GPaco* + KD-T & 36.7 & 48.6 & 37.7 & 42.2 \\
    GPaco* + KD-I & 37.0 & 48.5 & 37.3 & 42.2 \\
    GPaco* + KD-I-T & 37.3 & \textbf{48.9} & \textbf{38.9} & \textbf{42.8} \\
    \midrule
    Vision-only teacher model & \textcolor{gray}{51.6} & \textcolor{gray}{48.5} & \textcolor{gray}{36.2} & \textcolor{gray}{47.2} \\
    \bottomrule
  \end{tabular}
  \caption{Comparisons on Places-LT with ResNet-152. (* indicates our reproduced result.)}
  \label{table:main_table_placelt}
\end{table}

%% file: Tables/main_diffarch.tex
\begin{table}[b!]
  \begin{tabular}{p{2.6cm}p{0.5cm}p{0.5cm}p{0.5cm}p{0.5cm}}
    \toprule
    Method & Many & Med & Few & Avg \\
    \midrule
    Res-18 & 61.8 & 48.0 & 33.4 & 51.3 \\
    Res-18 + RISE & 63.1 & 50.8 & 35.8 & 53.5 \\
    Res-18 + KD-I & 63.0 & 50.6 & 35.5 & 53.3 \\
    Res-18 + KD-I-T & \textbf{64.2} & \textbf{51.5} & \textbf{38.3} & \textbf{54.6}\\
    \midrule
    Res-34 & 64.7 & 51.2 & 35.7 &  54.3 \\
    Res-34 + RISE & 65.7 & 52.3 & 37.3 & 55.5 \\
    Res-34 + KD-I & 65.9 & 52.2 & 36.8 & 55.4 \\
    Res-34 + KD-I-T & \textbf{67.5} & \textbf{54.3} & \textbf{41.0} & \textbf{57.6} \\
    \midrule
    LiVT (ViT-B/16) & 73.6 & 56.4 & 41.0 & 60.9 \\
    LiVT + RISE & 76.9 & 60.9 & 39.1 & 64.1 \\
    LiVT + KD-I & \textbf{77.0} & 60.7 & 38.8 & 64.0 \\
    LiVT + KD-I-T & \textbf{77.0} & \textbf{62.1} & \textbf{41.9} & \textbf{65.1} \\
    \bottomrule
  \end{tabular}
  \centering
  \caption{Comparisons on ImageNet-LT with different network architectures.}
  \label{table:main_table_diffarch}
\end{table}

%% file: Tables/main_scaling_img.tex
\begin{figure*}[t]
  \centering
    \includegraphics[width=0.9\textwidth]{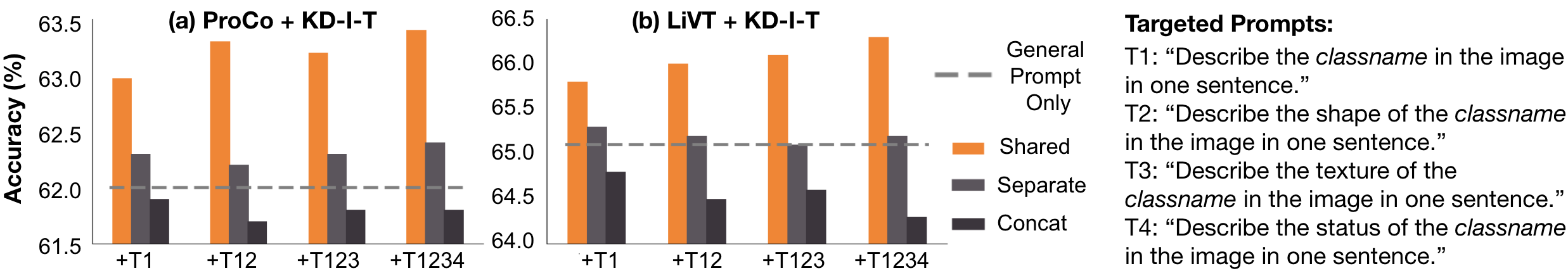}
  \caption{Benefits of adding more text supervision. The gray dash line indicates using only the captions generated by the General Prompt. +T1234 indicates training with captions generated by T1, T2, T3 and T4 Targeted Prompts.}
  \label{table:main_table_scale}
\end{figure*}

%% file: Tables/ablade_text.tex
\begin{table}[b]
  \centering
  \begin{tabular}{p{4cm}p{0.5cm}p{0.5cm}p{0.5cm}p{0.5cm}}
    \toprule
    Text Encoder & Many & Med & Few & Avg \\
    \midrule
    ProCo & 68.2 & 55.1 & 38.1 & 57.8 \\
    \midrule
    CLIP ViT-B/16 & \textbf{70.6} & 58.0 & 
    \textbf{42.1} & \textbf{60.7} \\
    CLIP ViT-L/14 & \textbf{70.6} & \textbf{58.1} & 40.4 & 60.5 \\
    Open-CLIP ViT-G/14 & 70.5 & 57.9 & 40.6 & 60.4 \\
    \midrule
    MPNet~\cite{song2020mpnet} & 70.4 & 57.9 & 40.2 & 60.3 \\
    \midrule
    CLIP ViT-B/16-FT  & 68.9 & 55.4 & 38.5 & 58.3 \\
    \bottomrule
  \end{tabular}
  \caption{Ablation study on the text encoder.}
  \label{abl:text}
\end{table}

%% file: Tables/ablade_naive.tex
\begin{table}[t]
  \centering
  \begin{tabular}{p{4cm}p{0.5cm}p{0.5cm}p{0.5cm}p{0.5cm}}
    \toprule
    Method & Many & Med & Few & Avg \\
    \midrule
    ProCo & 68.2 & 55.1 & 38.1 & 57.8 \\
    \midrule
    Classname & 68.3 & 55.2 & 38.7 & 58.0 \\
    VL-LTR~\cite{tian2022vl} & 68.3 & 55.2 & 38.7 & 58.2 \\
    Prompt-Short & \textbf{70.6} & 58.0 & \textbf{42.1} & \textbf{60.7}  \\
    Prompt-Long & 70.4 & 57.8 & 39.0 & 60.1 \\
    Keyword-Short & 69.8 & \textbf{58.1} & 40.4 & 60.2 \\
    Keyword-Long & 69.5 & \textbf{58.1} & 40.9 & 60.1 \\
    \bottomrule
  \end{tabular}
  \caption{Ablation study on types of text supervision.}
  \label{abl:prompt}
\end{table}

%% file: 05_ablation.tex
\section{Ablation study}
\label{abla}
In this section, we investigate the significance of each component in our approach. We begin by evaluating our approach with various text encoders. Next, we examine how our approach is influenced by different text supervision generated using different prompts. Then, we assess whether our approach is sensitive to different off-the-shelf VLMs. Finally, we analyze the performance of various text adaptors. For simplicity, we use ImageNet-LT as the evaluation benchmark and we pick ResNet-50 as the backbone and adopt the ProCo~\cite{du2024probabilistic} training procedure as the baseline.

\subsection{Choice of text encoders}
\label{abla:text}
In Table~\ref{abl:text}, we evaluate our approach using features generated from the same text dataset but with different text encoders. Our results indicate that the approach is not sensitive to the size of the encoder used. Notably, we observe a similar performance improvement with MPNet~\cite{song2020mpnet}, a text encoder trained solely on text supervision. This further supports the notion that text feature supervision offers complementary benefits when combined with image feature supervision. Additionally, we experimented with adding a text encoder (CLIP ViT-B/16) and fine-tuning it during training. Similar to CLIP, we supervised the text encoder with a contrastive loss based on their corresponding image features. However, we found that fine-tuning the text encoder during training, referred to as CLIP ViT-B/16-FT in Table~\ref{abl:text}, significantly decreased the distillation performance. Our intuition is that fine-tuning on a much smaller dataset (115.8k examples) compromised the generalization capability of the original text encoder.

\input{Tables/ablade_vlm}
\input{Tables/ablade_head}

\subsection{Different text datasets}
\label{abla:prompt}
In Table~\ref{abl:prompt}, we test our approach with different text supervision. 
``Classname'' represents text supervision generated from the label of the images, such as “a photo of a dog.” VL-LTR~\cite{tian2022vl} represents class-level text supervision, where each class is described by a few sentences sourced from Wikipedia. Prompt-Short is the general prompt illustrated in Figure~\ref{fig:overview}. We also experiment with the following prompts:
\begin{itemize}
    \item Prompt-Long: ``Please describe the image in detail.''
    \item Keyword-Short: ``Please describe the image with three keywords.''
    \item Keyword-Long: ``Please describe the image with ten keywords.''
\end{itemize}
As shown in the table, distillation from class-level text supervision offers marginal improvements over the baseline. However, applying text supervision from the VLM for each image results in significant enhancements across various settings. Our observations indicate that longer descriptions may lead to poorer distillation results, as distinct object-level features might be given less emphasis. Based on these findings, Prompt-Short works best for our settings.

\subsection{Choice of VLMs}
\label{abla:vlm}
LLaVA-NeXT~\cite{liu2024llavanext} offers multiple VLMs trained with language models of varying sizes, ranging from 7B to 34B parameters. Using Prompt-Short, we generated three text datasets with these different VLM models. However, we observed minimal changes in our distillation task's performance, as shown in Table~\ref{abl:vlm}. We suspect this is due to the classification task's complexity and the fact that all VLMs were trained with the same vision encoder. This demonstrates that our text supervisions can be generated with relatively low computational cost.

\subsection{Choice of text adaptors}
\label{abla:ada}
Vision encoders vary in their feature embedding sizes, necessitating the use of an adaptor to facilitate feature learning. For the text adaptor, we experiment with three different settings: \textit{linear layer}, \textit{1xMLP} + \textit{linear layer}, and \textit{2xMLP} + \textit{linear layer}, where \textit{MLP} consists of a linear layer, batch normalization, and ReLU activation. As shown in Table~\ref{abl:head}, the distillation performance is significantly worse without nonlinearities. Adding too many parameters to the text adaptor also leads to overfitting issues during training. We find that using a single nonlinear transformation followed by a linear layer yields the best results for our experimental settings.

%% file: Tables/ablade_vlm.tex
\begin{table}[b]
  \centering
  \begin{tabular}{p{2.5cm}p{0.5cm}p{0.5cm}p{0.5cm}p{0.5cm}}
    \toprule
    Method & Many & Med & Few & Avg \\
    \midrule
    ProCo & 68.2 & 55.1 & 38.1 & 57.8 \\
    \midrule
    Mistral-7b & \textbf{70.6} & \textbf{58.0} & \textbf{42.1} & \textbf{60.7}  \\
    Vicuna-13b & 70.4 & 57.9 & 40.2 & 60.3 \\
    Hermes-Yi-34B & 70.5 & \textbf{58.0} & 41.0 & 60.5 \\
    \bottomrule
  \end{tabular}
  \caption{Ablation study on different VLMs.}
  \label{abl:vlm}
\end{table}

%% file: Tables/ablade_head.tex
\begin{table}[t]
  \centering
  \begin{tabular}{p{2.5cm}p{0.5cm}p{0.5cm}p{0.5cm}p{0.5cm}}
    \toprule
    Method & Many & Med & Few & Avg \\
    \midrule
    ProCo & 68.2 & 55.1 & 38.1 & 57.8 \\
    \midrule
    Linear layer & 68.9 & 55.6 & 40.0 & 58.6 \\
    1xMLP + Linear & \textbf{70.6} & \textbf{58.0} & \textbf{42.1} & \textbf{60.7}  \\
    2xMLP + Linear & 70.3 & \textbf{58.0} & 40.1 & 60.1 \\
    \bottomrule
  \end{tabular}
  \caption{Ablation study on different text adaptors.}
  \label{abl:head}
\end{table}

%% file: 06_conclusion.tex
\section{Conclusion and limitations}
\label{con}

In this paper, we introduce a knowledge distillation approach using an off-the-shelf vision-language model. We have validated the effectiveness of our method on datasets such as ImageNet-LT, iNaturalist 2018, and Places-LT. Text supervision consistently offer complementary benefits to image features, and notably, the distillation performance enhances as more text supervision is generated using additional prompts. However, there are limitations to our approach. It requires a VLM model to generate text supervision for each image, which can be costly and thus may not be accessible to all researchers. 

%% file: 08_appendix.tex
\section{Appendix}
\label{app}
\input{Tables/app_scale_res}
\input{Tables/app_scale}
In the appendix, we first show the implementation details for experiments involving ImageNet-LT and iNaturalist 2018 in Section~\ref{app:model}. 
We then show more scaling benefits with ViT vision encoder on more text supervision in Section~\ref{app:scale}. Additionally, we visualize the feature distribution for different text supervision and compare it with image feature distribution in Section~\ref{app:feat}. Finally, we show more example prompts and responses from the generated text dataset in Section~\ref{app:example}.

\subsection{Implementation Details}
\label{app:model}
For all experiments, we adopt the original neural network architectures for ResNet-18, ResNet-34, ResNet-50 and ViT.
\paragraph{ImageNet-LT ProCo} We follow the same settings as described in~\cite{du2024probabilistic}. As mentioned in the paper, we change the batch size to 1024 and train for 180 epochs. We use two representation branches, which consists of a projection head with an output dimension of 2048 and a hidden layer dimension of 1024. The classification branch adopts a cosine classifier. We set the temperature parameter for the supervised contrastive learning to 0.07. RandAug is employed as a data augmentation strategy for the classification branch, and SimAug for the representation branch. We also assign equal loss weights to both branches. We use the same supervised contrastive loss as described in~\cite{du2024probabilistic}. 



\paragraph{ImageNet-LT LiVT} We use the same settings as described in~\cite{LiVT}. All models are trained with AdamW optimizer with $\beta$ as \{0.9, 0.95\}. We use the pretrained weights on ImageNet-LT provided by LiVT~\cite{LiVT}. The model is trained for 800 epochs with batch size 4096 and the mask ratio is 0.75. We finetune the model for 100 epochs for all our experiments. We train all models with RandAug (9, 0.5), mixup (0.8) and cutmix (1.0). All experiments set $\tau$ to 1. For the text adaptor, we directly apply it on the ViT embeddings and apply the classification head on the projected features after the adaptor. We find that this setup provides better results with LiVT training configurations.

\paragraph{iNaturalist 2018 ProCo} We change the batch size to 512 and keep the initial learning rate to 0.1. The learning rate decays by a cosine scheduler from 0.1 to 0 within 90 epochs of training. For other hyper-parameters, we follow the same settings as described in~\cite{du2024probabilistic}. Our reproduced results are similar to the results reported in \cite{du2024probabilistic} when trained for 90 epochs. All models are trained using SGD optimizer with momentum 0.9. 


\subsection{Scaling with ProCo and LiVT}
\label{app:scale}
As shown in Figure 3 of the main paper, we add the data sequentially and show the results for Many, Med and Few splits in Table~\ref{app:table:scale:proco} and Table~\ref{app:table:scale}. We can see that adding more data provides even more benefits for the Few split, which emphasizes the importance of applying our approach for long-tail visual recognition task.

\subsection{Feature Visualization}
\label{app:feat}
We visualize the feature embeddings with t-SNE in Figure~\ref{app:img:feat1}. Figure~\ref{app:fig:final1_img} shows the final feature embeddings of the ViT-B/16 used by the vision-only teacher model. Figure~\ref{app:fig:final1_nogt}, Figure~\ref{app:fig:final1_gt}, and Figure~\ref{app:fig:final1_cls} shows the text encoder embeddings for text description generated with General Prompt, Targeted Prompt (T1), and groundtruth classnames respectively. We use the text encoder of CLIP ViT-B/16. We randomly draw five images from the following classes: [`dragonfly', `leafhopper', `refrigerator', `iron', `brassiere', `overskirt', `killer-whale', `polecat', `chest armor', `pot']. Comparing Figure~\ref{app:fig:final1_img} and Figure~\ref{app:fig:final1_nogt}, we can see that the vision encoder provides a different feature distribution compared to the encoded text supervision. For example, in Figure~\ref{app:fig:final1_nogt}, features from `overskirt' are overlapping with features from `brassiere' while they are well-separated in Figure~\ref{app:fig:final1_img}. Also, features from `refrigerator' are wide spread in Figure~\ref{app:fig:final1_img} while they are grouped more tightly in Figure~\ref{app:fig:final1_nogt}. Additionally, feature clusters in Figure~\ref{app:fig:final1_cls} are uniformly distributed in the feature space while there are some biases of the feature cluster distribution in Figure~\ref{app:fig:final1_gt}. For example, `chest armor' and `pot' are closer in the feature space.

\subsection{Dataset examples}
\label{app:example}
In Figures 5-12, we show example prompts and responses for different images in ImageNet-LT. Based on the qualitative studies, it is evident that the VLM generates responses that are both reasonable and accurate to each question. Furthermore, the VLM can identify common visual features, including color, shape, and texture.

%% file: Tables/app_scale_res.tex
\begin{table*}[t!]
  \centering
  \begin{tabular}{p{6.0cm}p{0.7cm}p{0.7cm}p{0.7cm}p{0.7cm}}
    \toprule
    Method & Many & Med & Few & Avg \\
    \midrule
    ProCo + General Prompt(G) & 71.0 & 60.1 & 43.2 & 62.0 \\
    ProCo + G \& Targeted Prompts-(T1) & 71.5 & 60.8 & 46.7 & 63.0 \\
    ProCo + G \& Targeted Prompts-(T12) & 71.6 & 61.0 & 47.9 & 63.3 \\
    ProCo + G \& Targeted Prompts-(T123) & 71.6 & 60.9 & 47.5 & 63.2 \\
    ProCo + G \& Targeted Prompts-(T1234) & 71.7 & 61.1 & 48.0 & 63.4 \\
    \bottomrule
  \end{tabular}
  \caption{Scaling with ProCo.}
  \label{app:table:scale:proco}
\end{table*}

%% file: Tables/app_scale.tex
\begin{table*}[t!]
  \centering
  \begin{tabular}{p{6.0cm}p{0.7cm}p{0.7cm}p{0.7cm}p{0.7cm}}
    \toprule
    Method & Many & Med & Few & Avg \\
    \midrule
    LiVT + General Prompt(G) & 77.0 & 62.1 & 41.9 & 65.1 \\
    LiVT + G \& Targeted Prompts-(T1) & 77.2 & 62.6 & 44.8 & 65.8 \\
    LiVT + G \& Targeted Prompts-(T12) & 77.4 & 62.8 & 45.0 & 66.0 \\
    LiVT + G \& Targeted Prompts-(T123) & 77.5 & 62.9 & 45.1 & 66.1 \\
    LiVT + G \& Targeted Prompts-(T1234) & 77.5 & 63.1 & 45.9 & 66.3 \\
    \bottomrule
  \end{tabular}
  \caption{Scaling with LiVT.}
  \label{app:table:scale}
\end{table*}